\documentclass[10pt,twocolumn,letterpaper]{article}

\usepackage[pagenumbers]{cvpr}

\usepackage{xspace}
\newcommand{\method}{PinGround\xspace}

\usepackage{booktabs}
\usepackage{multirow}
\usepackage{amsmath,amssymb}
\usepackage{tikz}
\usetikzlibrary{positioning,arrows.meta,fit,calc,backgrounds}

\newcommand{\none}{\texttt{None}\xspace}

\definecolor{cvprblue}{rgb}{0.21,0.49,0.74}
\usepackage[pagebackref,breaklinks,colorlinks,allcolors=cvprblue]{hyperref}

\def\confName{CVPR}
\def\confYear{2027}

\title{PinGround: Pinning Queries to Points\\for Parallel Grounding in Vision-Language Models}

\author{
  \textbf{Shuxuan Li}$^{1,2}$ \quad
  \textbf{Jialiang Kang}$^{2}$ \quad
  \textbf{Yanghao Zhou}$^{2}$ \\
  \textbf{Ziye Zhu}$^{2,3}$ \quad
  \textbf{Zhilin Zhao}$^{1}$\thanks{Corresponding authors.} \quad
  \textbf{Xiao Yang}$^{2}$\footnotemark[1] \\
  $^{1}$Sun Yat-sen University \quad
  $^{2}$Canva Research \quad
  $^{3}$Peking University
}

\begin{document}
\maketitle
\begin{abstract}
Vision-language models (VLMs) can localize through their text interface, grounding open-vocabulary objects, complex referring expressions, text lines and layout regions with strong semantic understanding.
They write the boxes one after another as a single sequence, although the instances of a query, unlike the words of a sentence, form an unordered set with no sequential dependence.
The number of decoding steps therefore grows with the number of instances, which makes dense scenes the most expensive.
Decoding the instances in parallel avoids this growth, but streams that cannot see each other duplicate some instances and miss others.
We propose \method, which writes a single point into each query as a spatial anchor and asks for the matching instance nearest to that point within a local radius, or \none.
Given the image, the anchored queries are independent, so they are decoded in parallel as leaves of a prefix tree that shares the image and the query, and the model is trained on the same tree.
Coarse-to-fine anchors refine only the regions where instances are found, duplicate suppression keeps one box per instance, and set-level reinforcement learning optimizes the set F1.
Only the prompt and the attention mask change, not the architecture.
On three different VLMs and 12 benchmarks covering detection, OCR, layout and referring, \method is at most 5.5 F1 below sequential decoding trained on the same data, while its speedup grows with instance density and reaches 21.9$\times$ on dense scenes.
\end{abstract}

\section{Introduction}
\label{sec:intro}

\begin{figure}[t]
\centering
\includegraphics[width=\linewidth]{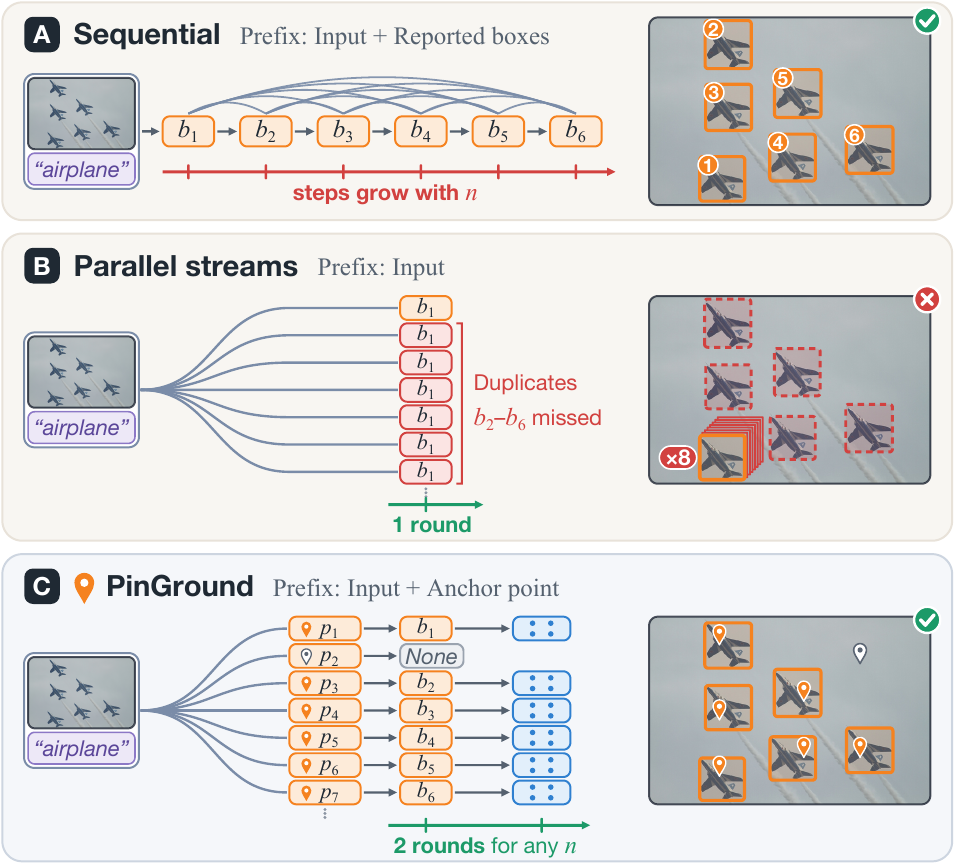}
\caption{Actual outputs of decoding all instances of a query on LocateAnything-3B. (A) Sequential decoding lists the boxes one by one, so its steps grow with $n$. (B) Parallel streams with identical inputs repeat one airplane and miss the rest. (C) \method gives each stream an anchor point and needs two rounds for any $n$.}
\label{fig:teaser}
\end{figure}

Computer vision is moving from task-specific models toward unified models that solve many tasks through one interface~\cite{chen2022pix2seqv2,wang2022ofa,lu2023unifiedio,xiao2024florence2,wang2023visionllm}.
Vision-language models (VLMs) bring this trend to localization.
Given an image and a text query, a single VLM returns the boxes of objects, text lines, layout regions or the referent of a free-form expression, writing each coordinate either as a number in ordinary text~\cite{chen2023shikra,wang2024qwen2vl,bai2025qwen3vl} or as a dedicated coordinate token~\cite{chen2022pix2seq,peng2023kosmos2,jiang2025rexomni}.
Open-vocabulary detectors such as Grounding DINO~\cite{liu2024groundingdino} also take text prompts and predict all boxes in parallel from a fixed set of object queries, but they encode the prompt with a relatively shallow text encoder such as BERT~\cite{devlin2019bert}, which suits category names and short phrases but not complex descriptions~\cite{jiang2025rexomni,jiang2025humanref}.
A VLM reads the query with a full language model, which makes it a strong candidate for a general-purpose localizer.

The language model of a VLM generates the answer by next-token prediction, so the boxes come out one after another as a single sequence (\cref{fig:teaser}A).
This sequential form is inherited from the language interface, not required by grounding.
Unlike the words of a sentence, the instances of a query carry no sequential dependence.
Any permutation of them is an equally correct answer, and each of them can be located from the image and the query alone.
The number of decoding steps, that is, sequential forward passes of the language model, nevertheless grows with the number of instances.
Rex-Omni reports that generation time grows approximately linearly with the number of predicted boxes and exceeds 16 seconds for images with more than 400 objects~\cite{jiang2025rexomni}, so dense scenes such as crowds, aerial images, documents and scene text take the longest to decode.
Existing accelerations make each instance cheaper but keep the list.
Rex-Omni writes each coordinate as one dedicated token instead of a string of digits~\cite{jiang2025rexomni}, LocateAnything predicts the four coordinates of a box in one step by multi-token prediction~\cite{wang2026locateanything}, and GroundAnything decodes fixed-length blocks of the output in parallel by masked diffusion~\cite{yu2026groundanything}, yet in all three the number of forward passes still grows with the number of instances.

This suggests decoding all instances of a query in parallel.
In sequential decoding, however, the earlier boxes show which instances have already been reported.
LocateAnything observes that seeing them lets the model avoid duplicate and missing boxes, and that parallelizing the output directly during training risks disrupting this causal process~\cite{wang2026locateanything}.
Parallel streams that see only the image and the query lose this signal, so they duplicate some instances and miss others (\cref{fig:teaser}B).
A parallel stream therefore needs to be told which instance to report.
Instances of a query differ in where they are, so we specify the instance by a location.
We write a point into the query as an anchor and ask for the box of the matching instance whose center is nearest to it, or for \none when no such center lies within a radius given by the anchor spacing (\cref{fig:teaser}C).
Each anchor thereby covers a local neighborhood, yet attends to the full image.
Given the image, the anchored queries are independent and can all be decoded in parallel.

Anchors lie on a coarse and a fine grid, and only the cells in which a coarse anchor finds an instance are refined, so the computation adapts to the density of the scene.
Neighboring anchors, as well as different queries, can still report the same instance, and duplicate suppression keeps one box per instance.
Parallel decoding relies on each anchor keeping to its own neighborhood, so that the anchors together miss and duplicate as few instances as possible.
A released VLM such as Qwen3-VL-4B does not keep to this rule, even when the rule is stated in the prompt.
We therefore adapt the VLM with a light post-training on 47k (image, query) pairs.
Training and decoding share a prefix-tree layout, in which the image is the root, each query is a branch and each anchor is a leaf holding only its point and its answer.
The attention mask confines every leaf to its own path, while the image and each query are encoded once for all their anchors.
Supervised fine-tuning on this layout scores each anchor in isolation, whereas the output is evaluated as a set.
Training therefore ends with set-level reinforcement learning (RL), which samples complete outputs of the decoder, scores each by set F1 and credits every anchor with its contribution.

We apply \method to three different VLMs, LocateAnything-3B~\cite{wang2026locateanything}, Qwen3-VL-4B~\cite{bai2025qwen3vl} and GroundAnything-VLM~\cite{yu2026groundanything}, and evaluate it on 12 benchmarks covering object detection, OCR, layout analysis and referring.
Each base is compared with its official decoder and with a sequential baseline fine-tuned on the same data.
Averaged over the 12 benchmarks, with each decoder in its default mode, \method trails this baseline by 1.3 F1 on LocateAnything, exceeds it by 4.3 F1 on Qwen3-VL and trails it by 1.0 F1 on GroundAnything.
Its speedup grows with instance density, from 3.1, 13.1 and 8.7 times on average over the 12 benchmarks to 9.3, 15.2 and 21.9 times on the dense detection benchmark Dense200, respectively.
Our contributions are as follows.
\begin{itemize}
\item We propose \method, a hierarchical spatial-anchor strategy that turns VLM grounding into independent queries decoded in parallel without architectural changes.
\item We adapt VLMs to this strategy with a light post-training that shares one prefix tree between training and decoding and ends with set-level reinforcement learning.
\item On three VLMs and 12 benchmarks, \method reaches F1 comparable to sequential decoding and is an order of magnitude faster on dense scenes.
\end{itemize}

\section{Related Work}
\label{sec:related}

\begin{figure*}[t]
\centering
\includegraphics[width=\textwidth,trim={9pt 0pt 8pt 0pt},clip]{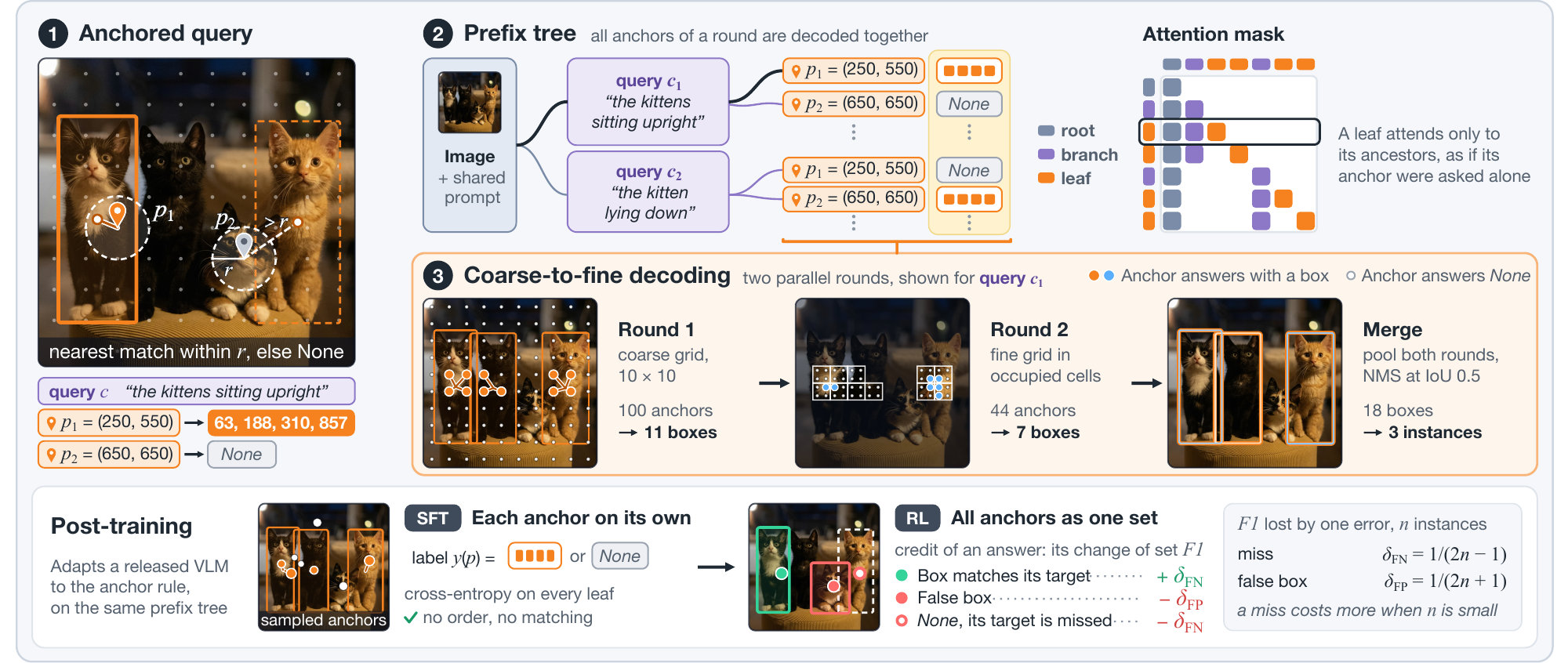}
\caption{Overview of \method. (1) An anchor asks for the matching instance nearest to it within radius $r$, or \none. (2) The anchored queries share one prefix tree in which each leaf attends only to its own path, so all anchors of a round are decoded in parallel. (3) A coarse $10\times10$ round finds the occupied cells, a fine round refines them, and NMS merges both rounds. Post-training teaches the anchor rule by SFT on every leaf and by RL that credits each anchor with its change of set F1.}
\label{fig:method}
\end{figure*}

\paragraph{Grounding with VLMs.}
Pix2Seq casts detection as generating a sequence of coordinate tokens and already notes that the order of the objects and the end of the sequence are hard to decide~\cite{chen2022pix2seq}.
This language interface now covers referring, grounded dialogue and unified vision tasks~\cite{peng2023kosmos2,chen2023shikra,you2024ferret,wang2023visionllm,wu2024visionllmv2,xiao2024florence2}, and general VLMs such as Qwen3-VL list many categories and instances in one answer~\cite{wang2024qwen2vl,bai2025qwen25vl,bai2025qwen3vl}.
Grounding VLMs trained at scale on detection, OCR, layout and referring data, such as Rex-Omni~\cite{jiang2025rexomni}, whose evaluation suite we adopt, LocateAnything~\cite{wang2026locateanything} and GroundAnything~\cite{yu2026groundanything}, handle all these tasks in one model.
To shorten the decoding of their growing answer, LocateAnything predicts the four coordinates of a box in one step and GroundAnything decodes blocks of tokens in parallel by masked diffusion.
Like general accelerations of LLM decoding~\cite{stern2018blockwise,leviathan2023speculative,cai2024medusa,gloeckle2024mtp}, both make each instance cheaper, but the instances still follow each other in one list.
\method instead changes the query, so that the number of decoding steps no longer grows with the number of instances.

\paragraph{Parallel queries for localization.}
Dedicated detectors regress all boxes in parallel, as DETR does from a set of learned queries with one-to-one matching~\cite{carion2020detr}.
Anchor DETR ties each query to an anchor point~\cite{wang2022anchordetr}, and open-set detectors such as GLIP, OWL-ViT and Grounding DINO read the prompt with a text encoder~\cite{li2022glip,minderer2022owlvit,liu2024groundingdino}.
GiT and UFO bring this parallelism to text outputs, and UFO applies it to a pretrained VLM~\cite{wang2024git,tang2026ufo}.
Each point of a fixed grid receives the local image feature as an extra input token and writes the category and box of the object there, or marks the point as empty.
Each point thus asks what lies there, which serves detection and segmentation over a category list given in the prompt, while a referring expression is still decoded as a single box in sequence.
\method instead asks where the target lies near each point and writes the target and the anchor coordinates into an ordinary prompt of the VLM, so one protocol decodes open-vocabulary detection, OCR, layout and multi-instance referring in parallel on different VLMs.
PET localizes people in crowds with point queries on a sparse grid and splits a query into four finer ones where a learned splitter judges the region dense~\cite{liu2023pet}, as \method refines a coarse anchor where it finds an instance.

\section{Method}
\label{sec:method}

\subsection{Anchored Queries and Prefix Tree}
\label{sec:queries}

Given an image $I$ and a text query $c$, which may be a category name, a text or layout class or a referring expression, grounding asks for the set $\mathcal{B}=\{b_1,\dots,b_n\}$ of boxes of all instances that match $c$.
Each box $b=(x_1,y_1,x_2,y_2)$ is given by its top-left and bottom-right corners, normalized by the image size to a scale of 1000.
A sequential VLM generates $\mathcal{B}$ as a list, so the number of its decoding steps grows with $n$.

\paragraph{Anchored queries.}
\method replaces the list by a set of anchored queries.
An anchor is a point $p$ written into the query, and its answer is the box of the matching instance whose center is nearest to $p$ if that center lies within a radius $r$, and \none otherwise.
Writing $\mathrm{ctr}(\cdot)$ for the center of a box and $j^\star=\arg\min_j \lVert \mathrm{ctr}(b_j)-p\rVert$ for the nearest instance, the answer is
\begin{equation}
y(p)=\begin{cases} b_{j^\star} & \text{if } \lVert \mathrm{ctr}(b_{j^\star})-p\rVert \le r,\\ \none & \text{otherwise.}\end{cases}
\label{eq:label}
\end{equation}
Each answer depends only on the image, the query and the anchor, so the anchored queries are independent and can be decoded in parallel, and the same rule supervises every anchor without an order of the instances or a matching step.
Together, the anchors divide the image among themselves, each responsible for the instances near it.
In normalized coordinates, we use two levels of anchors, the centers of a coarse ${10\times10}$ grid with spacing $100$ and of a fine ${20\times20}$ grid with spacing $50$.
Setting the radius of each level to its spacing makes neighboring anchors overlap, so that instances near cell borders stay within reach.

\paragraph{Prefix tree.}
The anchored queries of an image share the image and most of the prompt, and differ only in the query and the point.
We therefore arrange them as a prefix tree (\cref{fig:method}).
The root holds the image and the part of the prompt common to all queries.
Each query is a branch that holds the rest of the prompt with $c$ and the opening tokens of the response, which are the same for all its anchors.
Each anchor is a leaf that holds only its point and its answer, since the point is the only input that depends on the anchor.
The attention mask lets every token attend only to its own path from the root, and all leaves of a branch start at the same position index.
Each anchor thus sees exactly the context it would see if it were asked alone, while the image and the query are encoded once for all their anchors.
Training uses the same tree, with one (image, query) pair per row and its sampled anchors as leaves under the same mask and position indices, so training and decoding compute the same function.

\subsection{Coarse-to-Fine Decoding}
\label{sec:decoding}

Decoding takes one round per anchor level.
After a single prefill of the root, the first round computes each query branch once, together with the coarse anchors of all queries, which share these branches.
Each coarse cell whose anchor does not answer \none is then refined into its four fine anchors, which are decoded in the second round on the same root and branches.
The fine level is needed because each anchor reports at most one instance, so coarse anchors alone would recover at most one instance per cell, too few for small or densely packed instances such as text lines, crowds and objects in aerial images.
Refining only the occupied cells raises this bound where instances are found and keeps empty regions cheap, so the computation adapts to the density of the scene.
All anchors of a round are decoded together in the native output format of the base, so the number of sequential forward passes does not grow with the number of instances.

The boxes of both levels are pooled, and the same instance can appear several times in this pool, because neighboring anchors overlap, the fine anchors of a refined cell report again the instance found by its coarse anchor, and the model follows the rule of \cref{eq:label} only approximately.
We therefore apply non-maximum suppression (NMS) within each query, keeping the most confident box of each instance with the confidence computed from the token probabilities of the answer, and then across the queries of an image, since different queries such as car and van can return the same object.
Both steps use the IoU threshold $0.5$ of the main evaluation metric and involve no learned component.

\subsection{Supervised Post-Training}
\label{sec:sft}

The parallel decoder of \method relies on every anchor following the rule of \cref{eq:label}, which grounding itself does not require.
Applied directly, even with the rule stated in the prompt, the released bases do not answer in this way, and the decoder reaches only 3.6 to 48.0 F1@0.5 on Dense200, against 63.7 to 83.7 after post-training.
We therefore post-train the VLM to learn the rule, first by supervised fine-tuning and then by the set-level RL of \cref{sec:rl}.

Each training pair consists of an image, a query and its annotated boxes, and \cref{eq:label} labels every anchor of both levels directly from the annotations.
For each pair we draw up to four anchors with a target, uniformly from both levels, and at each level as many \none anchors as anchors with a target.
Coarse \none anchors are drawn from all coarse anchors without a target, since the decoder queries the whole coarse grid, and cover both empty background and regions next to other objects.
Fine \none anchors are drawn only from the cells that the decoder may refine, whose coarse anchor has an annotated instance within its radius, so they lie next to instances of the same or of other queries.
Pairs whose query does not occur in the image contribute eight coarse \none anchors.

Given the image $I$, the query $c$ and an anchor $p$, the model is trained to generate the answer $y(p)$ of \cref{eq:label}, the four coordinates of $b_{j^\star}$ or \none, written as tokens in the native output format of the base, with the token-level cross-entropy
\begin{equation}
\mathcal{L}=-\sum_{p}\sum_{t}\log \pi_\theta\big(y_t(p)\mid I,c,p,y_{<t}(p)\big),
\label{eq:sft}
\end{equation}
where $y_t(p)$ is the $t$-th token of $y(p)$ and the sum runs over the sampled anchors of the prefix tree.
To preserve the native interface of the base and mitigate catastrophic forgetting, the training mixture also replays the same pairs in its native list format.

\subsection{Set-Level Reinforcement Learning}
\label{sec:rl}

\begin{figure}[t]
\centering
\includegraphics[width=\linewidth]{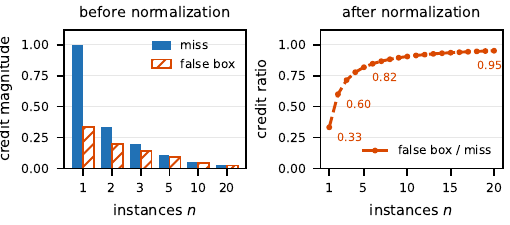}
\caption{Credit of a miss and of a false box versus the number of instances $n$. Left: before normalization, both credits shrink toward zero as $n$ grows. Right: after normalizing a pair whose largest credit is the miss, the miss receives $1$ and a false box $(2n{-}1)/(2n{+}1)$, which stays between $1/3$ and $1$.}
\label{fig:credit}
\end{figure}

\begin{table}[t]
\centering
\small
\caption{Credit of an anchor answer given its outcome in the matched output set, where $\delta_{\mathrm{FP}}$ and $\delta_{\mathrm{FN}}$ are the drops of F1 caused by one false box and by one miss in an image with $n$ instances. The box rows apply to anchors of both levels.}
\label{tab:credit}
\setlength{\tabcolsep}{3pt}
\begin{tabular}{@{}llr@{}}
\toprule
Answer & Outcome in the matched set & Credit \\
\midrule
box & matches its target & $+\delta_{\mathrm{FN}}$ \\
 & matches another anchor's target & $0$ \\
 & removed by suppression & $0$ \\
 & false box & $-\delta_{\mathrm{FP}}$ \\
 & false box, target missed & $-\delta_{\mathrm{FP}}-\delta_{\mathrm{FN}}$ \\
\midrule
fine \none & no target & $+\delta_{\mathrm{FP}}$ \\
 & target missed & $-\delta_{\mathrm{FN}}$ \\
 & target found by another anchor & $0$ \\
\midrule
coarse \none & $k$ instances in its cell missed & $-\min(k,3)\,\delta_{\mathrm{FN}}$ \\
\bottomrule
\end{tabular}
\end{table}

Supervised post-training judges each anchor answer on its own, whereas the set F1 judges the merged answers of all anchors of a query, so the per-anchor loss misjudges what an error costs in three ways.
First, several anchors report each instance without seeing each other's boxes, and an imprecise duplicate whose IoU with the kept box is below $0.5$ survives suppression as a whole false box, so with hundreds of anchors per query even a small rate of such answers adds several false boxes.
Second, a wrong coarse \none is a single wrong token for its anchor, but it prunes its cell and removes every instance inside from the output.
Third, the cost of an error depends on the number $n$ of instances in the image.
With TP true boxes, FP false boxes and FN missed instances, $\mathrm{F1}=2\,\mathrm{TP}/(2\,\mathrm{TP}+\mathrm{FP}+\mathrm{FN})$.
Starting from a perfect output with $\mathrm{TP}=n$ and $\mathrm{FP}=\mathrm{FN}=0$, one false box lowers F1 to $2n/(2n{+}1)$, a drop of $\delta_{\mathrm{FP}}=1/(2n{+}1)$, and one miss lowers it to $2(n{-}1)/(2n{-}1)$, a drop of $\delta_{\mathrm{FN}}=1/(2n{-}1)$.
A miss thus outweighs a false box in a sparse image, the two costs converge in a dense one, and on an empty image a single false box turns F1 from one into zero.

We therefore optimize the set F1 with RL.
For a pair $(I,c)$, the coarse-to-fine decoder runs $G=2$ times with sampling at temperature $0.7$, and each run yields an answer $\hat y(p)$ for every decoded anchor $p$ and an output set.
GRPO~\cite{shao2024deepseekmath} would give one reward to the whole output and share it among its hundreds of anchor answers.
Following difference rewards~\cite{wolpert2001optimal}, which MCR adds to the GRPO advantage of each box in a sequential answer~\cite{han2026mcr}, we instead credit each anchor with the change of F1 caused by its answer.
The target of an anchor is its nearest instance within the coarse radius, with box $y^\star(p)$, or \none if there is none.
As in the evaluation, the output set is matched one-to-one to the annotated boxes at IoU $0.5$, which determines whether each answer is a true box, a false box or a \none whose target is missed, and each role receives the corresponding change of F1 at the perfect set (\cref{tab:credit}).
A true box earns $\delta_{\mathrm{FN}}$ only if it matches the target of its anchor, so an anchor gains nothing by answering in place of its neighbors, a false box, including a duplicate that survives suppression, costs $\delta_{\mathrm{FP}}$, and a coarse \none is charged $\delta_{\mathrm{FN}}$ for each missed instance in its cell, clipped at three instances.
Since all credits of a pair scale with $1/n$ (\cref{fig:credit}, left), we divide them by their largest magnitude to obtain $a_p\in[-1,1]$, which puts all images on the same scale and keeps the ratio $\delta_{\mathrm{FP}}/\delta_{\mathrm{FN}}=(2n{-}1)/(2n{+}1)\in[1/3,1)$ within an image (\cref{fig:credit}, right).

An anchor with $a_p>0$ reinforces $\hat y(p)$, and an anchor with $a_p<0$ pushes $\hat y(p)$ down and learns $y^\star(p)$ in an added branch of the prefix tree.
Writing $\pi_\theta(y)$ for $\pi_\theta(y\mid I,c,p)$, the loss is
\begin{equation}
\begin{aligned}
\mathcal{L}_{\mathrm{RL}}={}&-\sum_{p} a_p\log \pi_\theta(\hat y(p))\\
&-\sum_{a_p<0}|a_p|\log \pi_\theta(y^\star(p))+\beta\,\mathrm{KL}(\pi_\theta\,\|\,\pi_{\mathrm{sft}}),
\end{aligned}
\label{eq:rl}
\end{equation}
where the sums run over the decoded anchors of both samples and the log-probability of an answer sums over its tokens as in \cref{eq:sft}.
The first term is the policy gradient with $a_p$ as advantage, and for $a_p<0$ its per-token cross-entropy is clipped at $4$, which stops the push-down once the probability of a token falls below $e^{-4}$.
The second term teaches every anchor with $a_p<0$ the answer for its target, and the KL term, taken over the full vocabulary at every supervised token with $\beta=0.5$, keeps the model close to the frozen supervised model $\pi_{\mathrm{sft}}$.
Rows of the supervised mixture of \cref{sec:sft}, including its native-format replay, are mixed in at half the amount of the RL data, and all pairs come from the supervised training set, so this stage adds no annotation.

\section{Experiments}
\label{sec:experiments}

\subsection{Setup}
\label{sec:setup}

\paragraph{Base models.}
We apply \method to three VLMs with different coordinate formats and decoding schemes.
LocateAnything-3B (LA)~\cite{wang2026locateanything} writes each coordinate as one of the tokens \texttt{<0>} to \texttt{<1000>} and has three decoding modes.
The fast mode predicts the four coordinates of a box in one step with a multi-token block, the slow mode writes them token by token, and the default hybrid mode falls back to token-by-token decoding for boxes whose block prediction is ambiguous.
Qwen3-VL-4B-Instruct~\cite{bai2025qwen3vl} writes coordinates as plain-text integers from $0$ to $1000$, and GroundAnything-VLM (GA-VLM)~\cite{yu2026groundanything} as the tokens \texttt{<0>} to \texttt{<999>}.
\method keeps the native coordinate format of each base for both the anchor and the answer (\cref{tab:templates}).
On LA, the fast, hybrid and slow modes of \method keep the block prediction of every anchor, re-decode the boxes flagged by the rule of LA, and re-decode every box answer, respectively, and we use the slow mode unless stated otherwise.
Re-decoding writes the four coordinates token by token for all selected anchors together, so it adds four forward passes per image however many boxes it covers.

\begin{table*}[t]
\centering
\footnotesize
\caption{F1@0.5 on the 12 box benchmarks and the mean F1@0.5, F1@mIoU and time per image. Official rows are the released models with their official decoders, and \mbox{-SFT} rows are these models fine-tuned on our pairs. UFO uses its multi-task model on the object sets and its referring model on the referring sets, and its means ($^*$) cover these six sets.}
\label{tab:main}
\setlength{\tabcolsep}{3.4pt}
\begin{tabular}{@{}lcccccccccccc|ccc@{}}
\toprule
 & \multicolumn{4}{c}{Object} & \multicolumn{4}{c}{Text} & \multicolumn{2}{c}{Layout} & \multicolumn{2}{c}{Referring} & \multicolumn{3}{c}{Mean} \\
\cmidrule(lr){2-5}\cmidrule(lr){6-9}\cmidrule(lr){10-11}\cmidrule(lr){12-13}\cmidrule(l){14-16}
Decoding & Dense & COCO & VisD. & LVIS & HierT. & TotalT. & SROIE & IC15 & DocL. & M6Doc & HRef & RefG & F1 & mIoU & s/img \\
\midrule
\multicolumn{16}{@{}l}{\textit{UFO (InternVL2.5-8B)}} \\
Official & 7.7 & 80.5 & 51.1 & 31.5 & -- & -- & -- & -- & -- & -- & 59.0 & 88.7 & 53.1$^*$ & 40.3$^*$ & 2.12$^*$ \\
\midrule
\multicolumn{16}{@{}l}{\textit{LocateAnything-3B}} \\
Official, hybrid & 75.7 & 79.1 & 61.8 & 68.7 & 72.5 & 76.1 & 81.6 & 57.0 & 90.1 & 90.6 & 83.2 & 88.6 & 77.1 & 60.2 & 2.78 \\
LA-SFT, fast & 64.0 & 77.1 & 53.2 & 69.1 & 70.4 & 70.6 & 61.8 & 40.7 & 88.0 & 78.3 & 81.9 & 87.7 & 70.2 & 55.4 & 1.89 \\
LA-SFT, hybrid & 84.2 & 79.7 & 62.3 & 71.7 & 81.5 & 73.5 & 88.0 & 47.9 & 90.4 & 88.9 & 83.9 & 88.2 & 78.4 & 61.8 & 2.23 \\
LA-SFT, slow & 85.6 & 79.2 & 64.4 & 72.5 & 83.4 & 74.1 & 93.0 & 49.7 & 92.8 & 90.4 & 83.4 & 87.8 & 79.7 & 62.1 & 6.27 \\
\method, fast & 70.3 & 71.7 & 48.4 & 63.1 & 65.6 & 67.3 & 72.5 & 44.5 & 85.7 & 81.8 & 82.0 & 86.4 & 69.9 & 55.2 & 0.56 \\
\method, hybrid & 74.0 & 72.6 & 52.0 & 64.2 & 71.1 & 69.5 & 79.1 & 47.3 & 89.7 & 85.4 & 82.6 & 86.7 & 72.9 & 57.8 & 0.66 \\
\method, slow & 81.0 & 74.7 & 62.5 & 66.7 & 78.8 & 71.3 & 89.6 & 50.3 & 91.5 & 89.0 & 82.7 & 86.7 & 77.1 & 61.4 & 0.71 \\
\midrule
\multicolumn{16}{@{}l}{\textit{Qwen3-VL-4B-Instruct}} \\
Official & 17.0 & 77.3 & 38.6 & 71.0 & 37.6 & 63.4 & 71.1 & 64.6 & 52.8 & 31.6 & 74.0 & 88.2 & 57.3 & 40.0 & 46.66 \\
Qwen3-VL-SFT & 43.8 & 76.5 & 43.8 & 69.4 & 66.9 & 74.0 & 89.6 & 56.2 & 83.7 & 76.8 & 84.1 & 88.4 & 71.1 & 53.6 & 27.04 \\
\method & 63.7 & 74.1 & 56.6 & 68.0 & 73.0 & 74.0 & 90.3 & 53.3 & 90.7 & 86.5 & 84.7 & 89.5 & 75.4 & 57.0 & 2.06 \\
\midrule
\multicolumn{16}{@{}l}{\textit{GroundAnything-4B}} \\
GA-VLM, official & 92.4 & 81.6 & 71.1 & 78.3 & 77.0 & 74.1 & 96.1 & 63.8 & 96.4 & 93.7 & 92.8 & 94.3 & 84.3 & 66.1 & 8.12 \\
GA-DLM, official & 91.6 & 82.2 & 64.7 & 80.8 & 73.2 & 84.8 & 96.5 & 83.1 & 97.2 & 93.8 & 93.3 & 99.6 & 86.7 & 70.1 & 2.08 \\
GA-VLM-SFT & 87.1 & 80.9 & 64.9 & 77.7 & 78.0 & 74.0 & 93.5 & 51.4 & 96.3 & 92.4 & 90.4 & 93.9 & 81.7 & 65.6 & 7.16 \\
\method & 83.7 & 78.3 & 66.4 & 74.4 & 77.5 & 77.9 & 90.6 & 54.8 & 94.0 & 91.1 & 87.9 & 91.4 & 80.7 & 64.5 & 0.82 \\
\bottomrule
\end{tabular}
\end{table*}

\paragraph{Training.}
All post-training uses the same 47{,}170 (image, query) pairs from the training sets of COCO and LVIS for objects~\cite{lin2014coco,gupta2019lvis}, HierText, ICDAR 2013, ICDAR 2015 and SROIE for text~\cite{long2022hiertext,karatzas2013icdar,karatzas2015icdar,huang2019sroie}, DocLayNet and M6Doc for layout~\cite{pfitzmann2022doclaynet,cheng2023m6doc}, and RefCOCOg and HumanRef for referring~\cite{mao2016refcocog,jiang2025humanref}.
The supervised stage runs for two epochs at learning rate $2\times10^{-6}$ on Qwen3-VL and GA-VLM.
On LA it runs for one epoch at $4\times10^{-6}$, and each answer is supervised both token by token and through the multi-token block of LA.
The RL stage makes one pass over the same pairs at the learning rate of the supervised stage (\cref{sec:impl}).

\paragraph{Baselines.}
For each base we report the released model with its official decoder and prompts.
We also fine-tune it on our pairs, with supervised learning only, to answer each query with the list of all its boxes in its native format, and name this sequential baseline by the suffix \mbox{-SFT}.
Since \method starts from the same weights, which were trained on far more data, comparing it with the \mbox{-SFT} baseline isolates the decoding protocol.
LA-SFT is decoded in all three modes of LA, and the released LA in its default hybrid mode.
Qwen3-VL-SFT writes compact lists {\small\texttt{[[x1,y1,x2,y2],...]}} instead of JSON to match the answer length of \method, and both Qwen3-VL baselines are queried once per category.
For GroundAnything we also report GA-DLM, the speculative diffusion decoder released with it, on its released inference engine.
We further report UFO~\cite{tang2026ufo} on InternVL2.5-8B~\cite{chen2024internvl25}, which decodes a fixed grid of points in parallel (\cref{sec:related}).
The category list in the prompt of its multi-task model holds the queried categories, except on COCO, its training set, where it holds all 80 classes as in its own evaluation.
We keep the boxes it predicts with a class score of at least $0.5$.
Its referring model decodes one box per expression.

\paragraph{Benchmarks and metrics.}
We evaluate on the 12 box benchmarks of the Rex-Omni evaluation suite~\cite{jiang2025rexomni}, covering object detection on Dense200, COCO~\cite{lin2014coco}, VisDrone~\cite{zhu2021visdrone} and LVIS~\cite{gupta2019lvis}, text detection on HierText~\cite{long2022hiertext}, TotalText~\cite{chng2017totaltext}, SROIE~\cite{huang2019sroie} and ICDAR2015~\cite{karatzas2015icdar}, layout analysis on DocLayNet~\cite{pfitzmann2022doclaynet} and M6Doc~\cite{cheng2023m6doc}, and referring on HumanRef~\cite{jiang2025humanref} and RefCOCOg~\cite{mao2016refcocog}.
The suite also contains pointing benchmarks, which we leave out because \method answers with boxes, whose centers trail native point decoding on average (\cref{sec:pointing}).
The composition of each test set is given in \cref{tab:benchmarks}.
A test sample is an image queried with the categories it contains, or one referring expression.
Following the released evaluation, the boxes of a sample are matched one-to-one to its ground truth greedily by IoU, precision and recall are averaged over samples, and F1 is computed from the averages.
We report F1 at IoU $0.5$ and its average over the IoU thresholds $0.5{:}0.95$ (F1@mIoU).
A \none counts as no prediction, and text detection is evaluated without recognition.

\paragraph{Latency.}
Accuracy and latency come from the same runs.
Each image is processed alone, with batch size one, by a single process on one NVIDIA H100 GPU, and the time per image covers preprocessing, vision encoding, prefill and decoding, measured with CUDA synchronization.
All models run in HF Transformers in eager mode, except GA-DLM, which runs on the SGLang-based engine released with it, and UFO, which runs in its released pipeline.
Sequential baselines and GA-DLM generate up to a maximum number of tokens per benchmark (\cref{sec:impl}), and an output that reaches it is evaluated and timed as it is.

\subsection{Main Results}
\label{sec:main}

\begin{figure}[t]
\centering
\includegraphics[width=\linewidth]{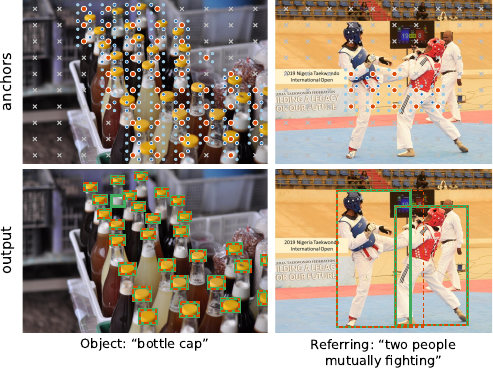}
\caption{Examples of \method on LA in the slow mode. The top row shows the anchors that report a box (orange coarse, blue refined) or \none (gray coarse, light blue refined), and the bottom row shows the output boxes (dashed) and the ground truth (green).}
\label{fig:qual}
\end{figure}

\begin{figure*}[t]
\centering
\includegraphics[width=\textwidth,trim={10 11 10 10},clip]{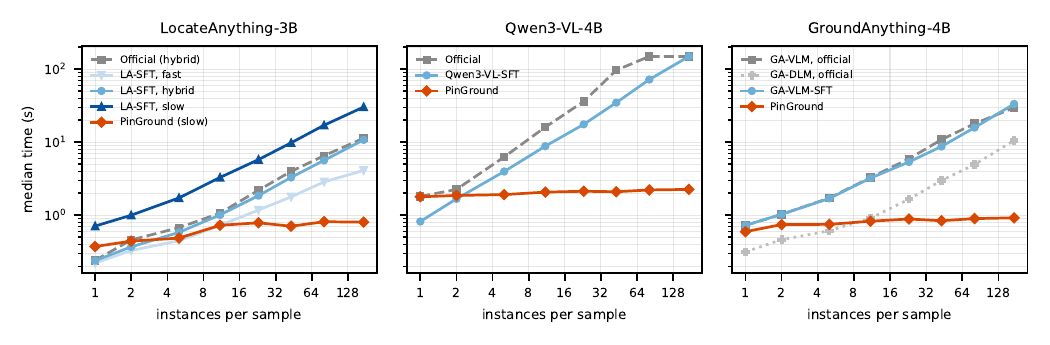}
\caption{Median time per sample versus the number of instances per sample on the 12 box benchmarks. Samples are grouped by instance count in powers of two, each point is the median of a group with at least 10 samples, and both axes are logarithmic.}
\label{fig:latency}
\end{figure*}

\Cref{tab:main} compares \method with the sequential baseline fine-tuned on the same pairs (\mbox{-SFT}), so that only the decoding protocol differs, and lists the released models with their official decoders for reference.
On LocateAnything, \method in its default slow mode reaches 77.1 F1 and 61.4 F1@mIoU at 0.71 seconds per image, against 78.4 and 61.8 at 2.23 seconds for LA-SFT in its default hybrid mode.
In matched modes, \method trails LA-SFT by 2.6 F1 in the slow mode and by 0.3 F1 in the fast mode, and it is 8.8 and 3.4 times faster.
Against the hybrid mode, the largest gaps are on COCO and LVIS, whereas \method is ahead on VisDrone, SROIE, ICDAR2015 and both layout sets.
\Cref{fig:qual} shows examples of its outputs.
On Qwen3-VL, \method is 4.3 F1 and 3.4 F1@mIoU above Qwen3-VL-SFT and takes 2.06 instead of 27.04 seconds per image.
The gains are largest on the dense Dense200 and VisDrone and on the two layout sets.
Both sequential baselines often fall into a loop of repeated or slowly shifting boxes until the token limit, Qwen3-VL-SFT on 7\% and the released model on 16\% of the samples of a benchmark on average, and over the remaining samples they still take 7.7 and 9.5 times longer than \method.

On GroundAnything, \method is 1.0 F1 below GA-VLM-SFT and takes 0.82 instead of 7.16 seconds per image, 8.7 times less.
The released GA-DLM is 2.5 times slower than \method, although it runs on its own inference engine, which decodes the same autoregressive model GA-VLM 1.1 to 1.2 times faster per token than our setup.
GA-DLM is also the most accurate row. The training data of GroundAnything is not released, and fine-tuning GA-VLM on our 47{,}170 pairs alone lowers its F1 from 84.3 to 81.7, so part of the gap to GA-DLM reflects our training data rather than the decoding protocol.

UFO also decodes its grid points in parallel, at 2.5 to 3.1 seconds per image on the object sets, almost independently of the number of instances.
Its released models were trained on COCO detection and on referring data, so it reaches 80.5 F1 on COCO but 7.7 on Dense200, whose categories are mostly outside COCO, and 31.5 on LVIS, and it decodes each referring expression as a single box, reaching 59.0 F1 on the multi-instance HumanRef.
On the text and layout sets, which it was not trained on, it finds no boxes, and we leave them out of its means.
These gaps reflect its training data as much as its protocol, and we list UFO for reference.

\subsection{Latency and Cost}
\label{sec:cost}

\begin{table}[t]
\centering
\footnotesize
\caption{Mean cost per image on Dense200 and per expression on RefCOCOg, with \method in the slow mode on LA. The table reports the forward passes of the language model including prefill (Fwds, rounded to integers), the TFLOPs of the matrix products computed from the processed tokens and attention lengths, the peak GPU memory in GiB (Mem) and the time in seconds (T).}
\label{tab:cost}
\setlength{\tabcolsep}{1.7pt}
\begin{tabular}{@{}l@{\hspace{2pt}}cccc@{\hspace{5pt}}cccc@{}}
\toprule
 & \multicolumn{4}{c}{Dense200} & \multicolumn{4}{c}{RefCOCOg} \\
\cmidrule(lr){2-5}\cmidrule(l){6-9}
Decoding & Fwds & TFLOPs & Mem & T & Fwds & TFLOPs & Mem & T \\
\midrule
\multicolumn{9}{@{}l}{\textit{LocateAnything-3B}} \\
LA-SFT, hybrid & 169 & 25.8 & 9.1 & 6.61 & 5 & 3.9 & 7.7 & 0.23 \\
\method, slow & 7 & 41.5 & 10.3 & 0.71 & 7 & 10.4 & 8.5 & 0.36 \\
\midrule
\multicolumn{9}{@{}l}{\textit{Qwen3-VL-4B-Instruct}} \\
Qwen3-VL-SFT & 885 & 15.5 & 8.6 & 31.07 & 20 & 3.5 & 8.4 & 0.79 \\
\method & 41 & 57.6 & 10.4 & 2.04 & 40 & 16.9 & 9.6 & 1.73 \\
\midrule
\multicolumn{9}{@{}l}{\textit{GroundAnything-4B}} \\
GA-VLM-SFT & 511 & 13.4 & 10.5 & 19.07 & 18 & 4.5 & 9.5 & 0.72 \\
\method & 13 & 40.8 & 11.0 & 0.87 & 13 & 13.1 & 9.6 & 0.59 \\
\bottomrule
\end{tabular}
\end{table}

\Cref{fig:latency} breaks the time down by the number of instances per sample.
From one to about 170 instances, the median time of \method grows by at most 2.2 times, against 19 to 182 times for the sequential decoders, GA-DLM included, since every additional box lengthens their output, whereas all anchors of a round of \method advance together in every forward pass.
With one or two instances, as in the referring sets, a sequential decoder needs about as many forward passes as our two rounds.
\method is then 1.2 to 1.4 times faster than the GA-VLM decoders, whose answers repeat the query before the box.
It is up to 1.7 times slower on LocateAnything, 1.9 times slower than GA-DLM, and 2.2 times slower on Qwen3-VL, where every digit is a separate token.

\Cref{tab:cost} counts the cost per image on the densest benchmark and on a single-target one.
On Dense200, \method needs 7 to 41 forward passes of the language model per image on average instead of 169 to 885, but 1.6 to 3.7 times the FLOPs and up to 1.8 GiB more peak memory, since every anchor of a round is computed, including the many that answer \none.
On RefCOCOg it needs 2.7 to 4.8 times the FLOPs and 0.7 to 2.0 times the passes.
Each anchor attends only to the shared prefix and its own few tokens, so the attention span of an answer stays fixed, whereas the $k$-th box of a list attends to all boxes before it.
\method thus trades extra computation for fewer sequential steps.
The trade pays off at batch size one, where a sequential decoder reads all weights to produce a single token and is bound by memory bandwidth~\cite{leviathan2023speculative}, whereas each pass of \method reads them once and advances the boxes of all instances together, so the speedup grows with the number of instances.

\subsection{Ablations}
\label{sec:ablation}

\begin{table}[t]
\centering
\footnotesize
\caption{Decoder components of \method on LocateAnything-3B in the slow mode, averaged over the 12 box benchmarks.}
\label{tab:abl_decoder}
\begin{tabular}{@{}lccccc@{}}
\toprule
Decoding & F1 & P & R & F1@mIoU & s/img \\
\midrule
Coarse grid only & 73.7 & 75.3 & 75.9 & 58.9 & 0.55 \\
Fine grid only & 76.3 & 74.1 & 81.0 & 60.6 & 1.14 \\
Coarse to fine & 77.1 & 72.6 & 84.9 & 61.4 & 0.71 \\
\bottomrule
\end{tabular}
\end{table}

\begin{table}[t]
\centering
\footnotesize
\caption{Post-training of \method on the three bases, averaged over the 12 box benchmarks.}
\label{tab:abl_post}
\setlength{\tabcolsep}{4.5pt}
\begin{tabular}{@{}llcccc@{}}
\toprule
Base & & F1 & P & R & F1@mIoU \\
\midrule
\multirow{2}{*}{\textit{LocateAnything-3B}} & SFT & 75.3 & 70.6 & 84.2 & 59.9 \\
 & + RL & 77.1 & 72.6 & 84.9 & 61.4 \\
\midrule
\multirow{2}{*}{\textit{Qwen3-VL-4B}} & SFT & 71.2 & 65.3 & 82.1 & 53.7 \\
 & + RL & 75.4 & 72.6 & 80.7 & 57.0 \\
\midrule
\multirow{2}{*}{\textit{GroundAnything-4B}} & SFT & 77.4 & 73.1 & 85.6 & 61.9 \\
 & + RL & 80.7 & 79.4 & 83.9 & 64.5 \\
\bottomrule
\end{tabular}
\end{table}

\paragraph{Coarse-to-fine decoding.}
\Cref{tab:abl_decoder} decodes LocateAnything with the coarse grid only, with the fine grid everywhere, and with the coarse-to-fine schedule.
The coarse grid alone is fastest but misses instances, with 9.0 points lower recall, because neighboring instances fall into one coarse cell and its anchor reports only one of them.
The fine grid everywhere, a single fixed grid like those of GiT and UFO, decodes four times as many anchors as the coarse grid and takes 1.14 instead of 0.71 seconds per image, yet its recall and F1 stay below those of coarse to fine.
Refining only the occupied cells recovers the instances of dense regions and adds 0.16 seconds per image to the coarse grid.

\paragraph{Post-training.}
RL raises F1 by 1.8, 4.2 and 3.3 points on LocateAnything, Qwen3-VL and GroundAnything (\cref{tab:abl_post}).
The gain comes from precision, which rises by 2.0, 7.3 and 6.3 points, while recall changes by less than two points.
As designed, RL mainly removes false boxes, the errors that the per-anchor supervised loss cannot see because it judges each answer alone (\cref{sec:rl}).

\section{Conclusion}
\label{sec:conclusion}

We have shown that a VLM need not generate the boxes of a query one after another.
A single point is enough to tell each parallel stream which instance to report, so the anchored queries are independent given the image.
\method decodes them as leaves of one prefix tree, in two rounds for any number of instances.
Only anchors that find an instance are refined.
SFT on the same prefix tree and set-level RL suffice to teach existing VLMs to answer such queries.
Without architectural changes, \method is at most 5.5 F1 below sequential decoding trained on the same data on three VLMs and up to 21.9 times faster on Dense200.
Its speedup grows with instance density, while queries with one or two targets gain little or even slow down, and its first-level anchors form a fixed grid that also spends computation on empty regions.
Adapting these anchors to each image and extending the protocol beyond the boxes studied here are left for future work.

{
    \small
    \bibliographystyle{ieeenat_fullname}
    \bibliography{main}
}

\clearpage
\maketitlesupplementary
\appendix
\counterwithin{table}{section}
\counterwithin{figure}{section}

\section{Implementation Details}
\label{sec:impl}

\paragraph{Training data.}
Of the 47{,}170 pairs, 12{,}000 are object queries on COCO images with COCO and LVIS categories~\cite{lin2014coco,gupta2019lvis}, 10{,}162 are text detection pairs from HierText, ICDAR 2013, ICDAR 2015 and SROIE~\cite{long2022hiertext,karatzas2013icdar,karatzas2015icdar,huang2019sroie}, 13{,}000 are layout queries from DocLayNet and M6Doc~\cite{pfitzmann2022doclaynet,cheng2023m6doc}, and 12{,}008 are referring expressions from RefCOCOg and HumanRef~\cite{mao2016refcocog,jiang2025humanref}.
The text detection, layout and HumanRef pairs are taken from the released training data of LA~\cite{wang2026locateanything}, and the object and RefCOCOg pairs are built from the original annotations.

\paragraph{Formats.}
\Cref{tab:templates} lists the anchor and answer formats on the three bases.
LA and GA-VLM have a native \none answer, whereas on Qwen3-VL \none is introduced by post-training.

\begin{table}[t]
\centering
\footnotesize
\caption{Anchor and answer formats of \method on the three bases. Each anchor is a point $(x,y)$ and each answer is one box $(x_1,y_1,x_2,y_2)$ or \none, written in the native coordinate format of the base. On LA the anchor follows \texttt{<ref>}$c$\texttt{</ref>}, which opens the response to the query $c$.}
\label{tab:templates}
\begin{tabular}{@{}l@{\hspace{5pt}}l@{}}
\toprule
\multicolumn{2}{@{}l}{LA: coordinate tokens \texttt{<0>} to \texttt{<1000>}}\\
Anchor & \texttt{<x><y>}\\
Answer & \texttt{<box><x1><y1><x2><y2></box>}, \texttt{<box>None</box>}\\
\midrule
\multicolumn{2}{@{}l}{Qwen3-VL: plain-text integers from $0$ to $1000$}\\
Anchor & \texttt{Point: (x, y).}\\
Answer & \texttt{[x1, y1, x2, y2]}, \texttt{None}\\
\midrule
\multicolumn{2}{@{}l}{GA-VLM: coordinate tokens \texttt{<0>} to \texttt{<999>}}\\
Anchor & \texttt{Point: <|box\char`\_start|><x><y><|box\char`\_end|>.}\\
Answer & \texttt{<|box\char`\_start|><x1><y1><x2><y2><|box\char`\_end|>},\\
 & \texttt{None}\\
\bottomrule
\end{tabular}
\end{table}

\paragraph{Supervised stage.}
On Qwen3-VL and GA-VLM each pair is trained as one sequence that holds the prefix tree of its anchors and one sequence of its native-format replay.
On these two bases the loss of the first answer token, which decides between a box and \none, is weighted by $3$.
A weight of $1$ gives a similar F1, and \cref{eq:sft} omits this weight for brevity.
On LA each pair likewise forms one prefix tree, with its anchors as leaves under the branch of its query, and these trees are packed together with the native replay sequences into sequences of up to 8{,}192 tokens.
LA predicts a box either token by token or with its multi-token block, which appends mask tokens to the input and predicts the whole answer at these positions in one forward pass.
Each anchor is supervised on both paths, with the loss of \cref{eq:sft} and the block loss of LA on the same answer.
The supervised stage of \method uses a cosine schedule after 50 warmup steps on all three bases.
The -SFT baselines are trained for two epochs at learning rate $2\times10^{-6}$ on all three bases, and on LA they also include the text pairs in its OCR-spotting format, as the replay of \method does.

\paragraph{RL stage.}
The rollouts are generated once from the supervised model with the coarse-to-fine decoder, including the suppression of duplicates within the query.
Qwen3-VL and GA-VLM sample each answer token at temperature $0.7$, and LA samples its block and the token-by-token re-decoding of every box at temperature $0.7$ with nucleus sampling at threshold $0.9$.
Credits smaller than $0.01$ in magnitude are discarded before the normalization.
Of the positively credited \none answers of a pair, at most the 16 least confident are kept.
Each credit weights the log-probability of its answer in \cref{eq:rl}, which covers every token of the answer on Qwen3-VL and GA-VLM and the block prediction on LA, that is, the coordinates, \none and the closing box token.
The learning rate follows a cosine schedule without warmup.

\paragraph{Decoding modes on LA.}
In the hybrid mode, \method re-decodes token by token the boxes flagged by the rule of the released LocateAnything code, that is, boxes whose block output is malformed or has a coordinate whose top-1 probability is below $0.9$ while the coordinate tokens among its four most probable ones spread over more than 60 bins.
The LA paper states $0.7$ and the five most probable tokens with a spread of 80~\cite{wang2026locateanything}, and we keep the released rule, which the official decoder also uses for the released LA and LA-SFT.
The slow mode re-decodes every box answer, and \none answers keep the prediction of the block in all modes.
The official decoder samples at temperature $0.7$ with nucleus sampling at threshold $0.9$ and a repetition penalty of $1.1$, so the official and LA-SFT rows vary between runs, and three runs of LA-SFT in the hybrid mode reach 84.2, 82.6 and 81.5 F1 on Dense200, the first of which is reported in \cref{tab:main}.
\method, the sequential Qwen3-VL and GroundAnything decoders and GA-DLM decode greedily.

\paragraph{Decoding limits and engines.}
The sequential LocateAnything decoders generate at most 8192 new tokens per output, and the sequential Qwen3-VL and GroundAnything decoders and GA-DLM at most 4096 on Dense200, VisDrone, HierText and SROIE, 2048 on the other object, text and layout sets and 1024 on the referring sets.
GA-DLM runs on its released SGLang-based engine, whose scheduler, attention kernels and KV cache are adapted to its decoder, in BF16 without CUDA graphs or overlap scheduling as released, and is timed per request.
With the same settings, this engine runs the autoregressive GA-VLM at 32.0 to 32.6 ms per generated token on 1{,}103 samples from five benchmarks, against 36.7 to 37.5 ms in our setup, both measured as the slope of the request time over the number of generated tokens.

\paragraph{Evaluation details.}
The evaluation code released with Rex-Omni and LocateAnything has two evaluators for box benchmarks~\cite{jiang2025rexomni,wang2026locateanything}.
A COCO-style evaluator scores COCO and LVIS, matching boxes within each category and averaging precision and recall over categories.
A general grounding evaluator scores the other benchmarks and averages precision and recall over samples.
It pools the boxes of all queried categories of a sample before matching, so a box can also match ground truth of another category.
\Cref{tab:main} scores all 12 benchmarks with the general evaluator (\cref{sec:setup}), so that they share one protocol.

\Cref{tab:cat_protocol} also scores the answers of \cref{tab:main} with a category-aware protocol, which keeps the per-sample average but, on the object and layout benchmarks, matches a box only to ground truth of its own category.
\method queries each anchor with a single category and writes no names, and both Qwen3-VL baselines are queried once per category, so for these models the category of a box is that of its query.
The other models are queried with all categories together, so the category of a box is the name written with it.
A box whose name is not a queried category, including a truncated or misspelled name, counts as a false box, as in \cref{tab:main}.
COCO crowd regions are not counted as targets and boxes inside them are ignored, as are unmatched LVIS boxes of categories that are not exhaustively annotated or not verified in the image and boxes lying mostly in illegible ICDAR2015 regions.
Every box is first read as the rectangle spanned by its two corners, whatever their order, then zero-area boxes are dropped and a box repeated with the same coordinates and category counts once.

The means change by $-0.5$ to $+1.1$ F1 on the three bases, and the gaps between \method and the \mbox{-SFT} baselines by at most $0.7$ F1.
Individual benchmarks change more and in both directions, VisDrone and ICDAR2015 the most.
Category matching lowers the VisDrone F1, since boxes of close classes such as car and van or pedestrian and people no longer match each other, and ignoring illegible regions raises the ICDAR2015 F1 of all models.
The VisDrone F1 of \method drops more than that of the \mbox{-SFT} baselines.
The LocateAnything and GroundAnything models write the boxes of all queried categories in one sequence, so they choose the category of each box among all queried categories and see the boxes already written.
Each query of \method asks for a single category and is decoded in parallel with the others without seeing their answers, so the queries of close classes often return the same object, and NMS keeps only the box of the more confident query (\cref{sec:decoding}), whose class is not always the annotated one.
The Qwen3-VL baselines also answer each category in a separate query, but they apply no NMS, so when the queries of two classes return the same object, both boxes are kept and the one of the annotated class still matches.

\begin{table*}[t]
\centering
\footnotesize
\caption{F1@0.5 of the answers of \cref{tab:main} under the category-aware protocol, with the mean F1@0.5 of \cref{tab:main} in the last column. The means of UFO cover its six sets as in \cref{tab:main}.}
\label{tab:cat_protocol}
\setlength{\tabcolsep}{3.4pt}
\begin{tabular}{@{}lcccccccccccc|cc@{}}
\toprule
 & \multicolumn{4}{c}{Object} & \multicolumn{4}{c}{Text} & \multicolumn{2}{c}{Layout} & \multicolumn{2}{c}{Referring} & \multicolumn{2}{c}{Mean} \\
\cmidrule(lr){2-5}\cmidrule(lr){6-9}\cmidrule(lr){10-11}\cmidrule(lr){12-13}\cmidrule(l){14-15}
Decoding & Dense & COCO & VisD. & LVIS & HierT. & TotalT. & SROIE & IC15 & DocL. & M6Doc & HRef & RefG & F1 & \cref{tab:main} \\
\midrule
\multicolumn{15}{@{}l}{\textit{UFO (InternVL2.5-8B)}} \\
Official & 7.7 & 81.4 & 42.3 & 31.7 & -- & -- & -- & -- & -- & -- & 59.0 & 88.7 & 51.8$^*$ & 53.1$^*$ \\
\midrule
\multicolumn{15}{@{}l}{\textit{LocateAnything-3B}} \\
Official, hybrid & 75.8 & 79.6 & 53.2 & 69.5 & 72.6 & 76.1 & 81.7 & 60.6 & 88.4 & 89.5 & 83.2 & 88.6 & 76.6 & 77.1 \\
LA-SFT, fast & 64.3 & 77.9 & 46.8 & 70.6 & 70.7 & 70.6 & 62.5 & 48.8 & 87.0 & 77.5 & 81.9 & 87.7 & 70.5 & 70.2 \\
LA-SFT, hybrid & 84.3 & 80.3 & 53.5 & 73.2 & 81.6 & 73.5 & 88.0 & 60.5 & 88.9 & 87.7 & 83.9 & 88.2 & 78.6 & 78.4 \\
LA-SFT, slow & 85.6 & 79.9 & 54.5 & 73.9 & 83.4 & 74.1 & 93.1 & 62.8 & 91.0 & 89.0 & 83.4 & 87.8 & 79.9 & 79.7 \\
\method, fast & 70.4 & 72.5 & 39.9 & 65.6 & 66.0 & 67.4 & 72.9 & 55.9 & 84.5 & 81.1 & 82.0 & 86.4 & 70.4 & 69.9 \\
\method, hybrid & 74.1 & 73.3 & 42.7 & 66.5 & 71.4 & 69.6 & 79.4 & 60.5 & 88.1 & 84.3 & 82.6 & 86.7 & 73.3 & 72.9 \\
\method, slow & 81.1 & 75.1 & 50.6 & 68.6 & 78.8 & 71.3 & 89.6 & 65.2 & 89.9 & 87.3 & 82.7 & 86.7 & 77.2 & 77.1 \\
\midrule
\multicolumn{15}{@{}l}{\textit{Qwen3-VL-4B-Instruct}} \\
Official & 19.5 & 78.9 & 41.0 & 72.9 & 41.8 & 64.6 & 71.2 & 71.5 & 47.4 & 29.1 & 74.1 & 88.2 & 58.4 & 57.3 \\
Qwen3-VL-SFT & 44.0 & 77.5 & 39.6 & 71.7 & 67.5 & 74.1 & 90.0 & 68.8 & 82.1 & 75.1 & 84.1 & 88.4 & 71.9 & 71.1 \\
\method & 63.7 & 74.6 & 46.0 & 70.0 & 73.0 & 74.0 & 90.3 & 68.3 & 88.5 & 83.7 & 84.7 & 89.5 & 75.5 & 75.4 \\
\midrule
\multicolumn{15}{@{}l}{\textit{GroundAnything-4B}} \\
GA-VLM, official & 92.6 & 82.3 & 63.1 & 80.1 & 77.3 & 74.1 & 96.1 & 66.9 & 95.9 & 92.7 & 92.8 & 94.3 & 84.0 & 84.3 \\
GA-DLM, official & 91.6 & 83.2 & 56.9 & 83.0 & 76.2 & 85.0 & 96.6 & 90.3 & 96.6 & 93.0 & 93.3 & 99.6 & 87.1 & 86.7 \\
GA-VLM-SFT & 87.4 & 81.6 & 56.2 & 79.6 & 78.2 & 74.0 & 93.5 & 64.2 & 95.5 & 91.5 & 90.4 & 93.9 & 82.2 & 81.7 \\
\method & 83.7 & 78.5 & 53.3 & 75.9 & 77.5 & 77.9 & 90.6 & 70.0 & 92.4 & 89.3 & 87.9 & 91.4 & 80.7 & 80.7 \\
\bottomrule
\end{tabular}
\end{table*}

\paragraph{Cost.}
The TFLOPs of \cref{tab:cost} count the products with the weights of the language model and the vision encoder for every processed token and image patch, and the attention products of the language model for every computed query-key pair, including the pairs hidden by the mask of the prefix tree.
The output layer, the attention inside the vision encoder and the normalization and activation functions are not counted.
With these included and the output layer computed for the predicted tokens, \method needs 1.4 to 3.6 times the FLOPs of the sequential decoders on Dense200 and 2.7 to 4.8 times on RefCOCOg.
Mem is the peak GPU memory allocated while processing a sample, including the weights.

\section{Benchmarks}
\label{sec:suppl_bench}

The referring sets and the 1610 VisDrone test images are used in full, COCO, DocLayNet and M6Doc are represented by 1000 images drawn at random, LVIS by the first 1000 validation images that come from COCO val2017 and are not in our training data, and the other sets are used as released (\cref{tab:benchmarks}).
Each image is queried only with the categories it contains, so that every query has at least one instance.
Boxes are parsed from the answer and any other text is ignored, so a reply without boxes counts as no prediction.

Each text image is queried once with the class \emph{text line} on HierText and SROIE or \emph{word} on TotalText and ICDAR2015, and its answer is scored as text detection.
The released evaluation matches a text box only to ground truth with the same transcription, so we merge all text boxes of an image into this class and match them by IoU alone.
We leave out transcription because the anchored answers of a round are decoded together, so a round lasts as long as its longest answer.
A box is a short answer of nearly fixed length on every base, whereas the transcriptions of words and text lines differ widely in length.
Reading each detected box with a separate recognizer would suit \method better, but it adds a specialist model, whereas VLM grounding aims to solve these tasks with one model.

\begin{table}[t]
\centering
\footnotesize
\caption{Test samples of the 12 benchmarks: images for the first three groups and expressions for referring.}
\label{tab:benchmarks}
\begin{tabular}{@{}llrl@{}}
\toprule
Task & Benchmark & Samples & Selection\\
\midrule
Object & Dense200 & 200 & as released\\
 & COCO & 1000 & random from 4952\\
 & VisDrone & 1610 & full test set\\
 & LVIS & 1000 & COCO val2017, not in training\\
Text & HierText & 1723 & as released\\
 & TotalText & 300 & as released\\
 & SROIE & 360 & as released\\
 & ICDAR2015 & 496 & as released\\
Layout & DocLayNet & 1000 & random from 6480\\
 & M6Doc & 1000 & random from 2724\\
Referring & HumanRef & 4991 & full, without duplicates\\
 & RefCOCOg & 4889 & full validation split\\
\bottomrule
\end{tabular}
\end{table}

\section{Pointing Benchmarks}
\label{sec:pointing}

Following LocateAnything~\cite{wang2026locateanything}, we also evaluate pointing on Dense200, VisDrone~\cite{zhu2021visdrone}, COCO~\cite{lin2014coco}, HumanRef~\cite{jiang2025humanref} and RefCOCOg~\cite{mao2016refcocog}.
This section gives the protocol and the results and explains why \method answers with boxes.

\paragraph{Setup.}
We use all 200 images of Dense200 and all 4{,}988 distinct expressions of HumanRef.
On VisDrone, COCO and RefCOCOg we take the first 500 images, 1000 images and 1000 expressions of the released files, which have about as many instances per sample as the full sets.
The official and \mbox{-SFT} models of LocateAnything and GroundAnything and the official Qwen3-VL answer with points.
Qwen3-VL-SFT answers with boxes even when asked for points, so its boxes and those of \method are each read as their center.
The time per image in \cref{tab:pointing} is measured as in \cref{tab:main}.

We score the points with the pointing evaluator released with LocateAnything~\cite{wang2026locateanything}, which matches a point only to instances of its own category.
\method and both Qwen3-VL baselines are queried with one category at a time, and so are the LocateAnything models in their released pointing evaluation, so the category of a point is that of its query.
The GroundAnything models are queried with all categories together and write a name with each point, which gives its category.
A point matches an instance if it lies inside the mask of the instance, or inside its box in the few samples without masks.
Each instance is matched by at most one point, and points left unmatched count as false positives.
As for boxes, precision and recall are averaged over samples and F1 is computed from the averages.
On COCO, crowd regions are not counted as instances, and points inside them that match no instance are ignored, as in the category-aware protocol of \cref{sec:impl}.

\paragraph{Results.}
Averaged over the three bases, \method reaches 68.5 F1 with its box centers, against 70.9 for the \mbox{-SFT} baselines (\cref{tab:pointing}).
On LocateAnything and GroundAnything, where the \mbox{-SFT} baselines answer with points, \method trails LA-SFT by 3.5 F1 and GA-VLM-SFT by 7.9 F1.
The gap comes mainly from COCO and RefCOCOg, where \method trails LA-SFT by 8.7 and 7.4 F1 and GA-VLM-SFT by 15.0 and 16.4 F1, and on HumanRef it trails them by 0.8 and 12.9 F1.
On Dense200, \method trails LA-SFT by 2.5 F1 and leads GA-VLM-SFT by 6.5 F1, and on VisDrone it leads LA-SFT by 1.9 F1 and trails GA-VLM-SFT by 1.8 F1.
On Qwen3-VL, where \method and Qwen3-VL-SFT both answer with box centers, \method leads Qwen3-VL-SFT by 4.3 F1, with the largest leads of 14.5 F1 on Dense200 and 8.1 F1 on VisDrone.
These are also the two sets on which it leads Qwen3-VL-SFT most on boxes (\cref{tab:main}).
The official Qwen3-VL, which answers with points, leads Qwen3-VL-SFT and \method on COCO and RefCOCOg, where a box center often misses its instance, as counted below.
\method keeps its speed, at 0.62, 2.01 and 0.78 seconds per image against 1.74, 32.80 and 7.29 for the \mbox{-SFT} baselines.
GA-VLM-SFT takes much longer than GA-VLM because it often repeats earlier points, writing 143 points per image on VisDrone for 44 instances, against 39 for GA-VLM.

\begin{table}[t]
\centering
\footnotesize
\caption{F1 on the pointing benchmarks and the mean time per image over the five sets. The official and \mbox{-SFT} rows of LocateAnything and GroundAnything and the official row of Qwen3-VL answer with points, and the other rows answer with boxes read as their centers.}
\label{tab:pointing}
\setlength{\tabcolsep}{3.3pt}
\begin{tabular}{@{}lccccc|cc@{}}
\toprule
Decoding & Dense & VisD. & COCO & HRef & RefG & Mean & s/img \\
\midrule
\multicolumn{8}{@{}l}{\textit{LocateAnything-3B}} \\
Official, hybrid & 83.8 & 58.0 & 72.8 & 71.6 & 76.3 & 72.5 & 1.66 \\
LA-SFT, hybrid & 84.1 & 55.1 & 74.0 & 72.1 & 74.0 & 71.9 & 1.74 \\
\method, slow & 81.6 & 57.0 & 65.3 & 71.3 & 66.6 & 68.4 & 0.62 \\
\midrule
\multicolumn{8}{@{}l}{\textit{Qwen3-VL-4B-Instruct}} \\
Official & 36.9 & 37.1 & 74.0 & 57.0 & 77.4 & 56.5 & 53.82 \\
Qwen3-VL-SFT & 51.9 & 44.9 & 68.1 & 73.0 & 68.7 & 61.3 & 32.80 \\
\method & 66.4 & 53.0 & 65.7 & 73.3 & 69.4 & 65.6 & 2.01 \\
\midrule
\multicolumn{8}{@{}l}{\textit{GroundAnything-4B}} \\
GA-VLM, official & 89.9 & 69.1 & 84.7 & 91.3 & 88.7 & 84.7 & 3.97 \\
GA-DLM, official & 88.1 & 61.2 & 84.5 & 91.0 & 74.2 & 79.8 & 1.89 \\
GA-VLM-SFT & 75.9 & 61.1 & 83.8 & 88.7 & 87.6 & 79.4 & 7.29 \\
\method & 82.4 & 59.3 & 68.8 & 75.8 & 71.2 & 71.5 & 0.78 \\
\bottomrule
\end{tabular}
\end{table}

\paragraph{Box centers as points.}
The center of a box can miss its instance even when the box is correct.
We count such misses on the 1000 COCO images of the box benchmark (\cref{tab:benchmarks}) and on the 1000 RefCOCOg expressions of \cref{tab:pointing}.
Of the boxes of \method on LocateAnything that match a ground-truth box of their query at IoU $0.5$, 14.1\% on COCO and 24.4\% on RefCOCOg have their center outside the instance mask, and the boxes of Qwen3-VL-SFT give similar shares of 15.0\% and 23.1\%.
Most of these centers belong to instances that fill little of their box.
For \method, instances whose mask fills less than 40\% of their box make up 18\% of the matched instances on COCO and 31\% on RefCOCOg, but 58\% and 72\% of the centers outside the mask.
By category, the centers of \method on COCO miss 56\% of the matched dining tables, 29\% of the chairs and 12\% of the persons, the most frequent category.
A box center is thus a reliable point for compact instances, but for instances whose shape or occlusion leaves much of their box uncovered, such as chairs and dining tables, it often falls on the background or on another object.

\paragraph{Points as answers.}
Training each anchor to answer with a point inside its instance would avoid these misses, but it makes duplicates hard to remove.
Neighboring anchors overlap and are decoded without seeing each other's answers, so one instance can be reported by several anchors, and \method relies on NMS to keep only the most confident of their overlapping boxes (\cref{sec:decoding}).
Points can only be compared by distance, and the points given for one large instance can lie farther apart than two neighboring small instances, so any fixed distance either leaves several points on one large instance or merges neighboring small instances, and both lower F1.
We test this on LocateAnything with a variant of \method whose anchors answer pointing queries with points, trained by SFT on its pairs together with 4{,}000 pointing pairs from the training data of LocateAnything~\cite{wang2026locateanything}.
As in NMS, a point is dropped if it lies within a fixed distance of a more confident point already kept for the same query.
On the COCO and Dense200 images of \cref{tab:pointing}, a distance of 10 in normalized coordinates leaves the duplicates on large instances, and the precision on COCO is 8.8\%.
A distance of 150 merges neighboring instances on Dense200, where the recall falls to 23.4\%, while the precision on COCO is still 49.2\%.
None of the distances we tested from 10 to 150 suits both sets.
Duplicates are the main difficulty of parallel decoding.
A point is a single coordinate without area, so the IoU-based suppression used for boxes cannot be applied to it, and parallel pointing needs a new strategy to remove duplicates, which is beyond the scope of this work and left for future work.

\section{Comparison with UFO}
\label{sec:local_feat}

UFO~\cite{tang2026ufo} also decodes many point queries of an image in parallel on a VLM (\cref{sec:related}).
\Cref{sec:ufo_eval} details our evaluation of UFO, \cref{sec:ufo_diff} states where the two designs differ and \cref{sec:feat_anchor} tests within \method the local image features that UFO gives its points.

\begin{table}[t]
\centering
\footnotesize
\caption{F1 at IoU $0.5$ of the released UFO models, as in \cref{tab:main}, when only the boxes scored at least the score threshold of each column are kept. In grid mode all grid points are decoded in parallel over a category list in the prompt, which holds on COCO either the queried categories of the image or all 80 categories and on the referring sets the expression as its only category. The grounding head decodes one box per expression in sequence without a score, so its F1 does not depend on the threshold.}
\label{tab:ufo_sweep}
\setlength{\tabcolsep}{3pt}
\begin{tabular}{@{}lccccccc@{}}
\toprule
Score threshold & 0.1 & 0.2 & 0.3 & 0.4 & 0.5 & 0.6 & 0.7 \\
\midrule
\multicolumn{8}{@{}l}{\textit{Objects, multi-task model in grid mode}} \\
Dense200 & 9.0 & 8.9 & 8.6 & 8.3 & 7.7 & 7.0 & 6.1 \\
COCO, queried categories & 51.8 & 62.2 & 67.9 & 70.7 & 72.4 & 73.3 & 73.7 \\
COCO, all 80 categories & 70.5 & 76.1 & 78.4 & 79.8 & 80.5 & 80.6 & 80.8 \\
VisDrone & 41.1 & 46.8 & 49.5 & 50.7 & 51.1 & 50.5 & 49.0 \\
LVIS & 27.0 & 29.8 & 30.8 & 31.3 & 31.5 & 31.2 & 30.8 \\
\midrule
\multicolumn{8}{@{}l}{\textit{Referring, referring model in grid mode}} \\
HumanRef & 17.1 & 16.9 & 14.6 & 11.4 & 7.8 & 5.1 & 3.3 \\
RefCOCOg & 16.3 & 16.9 & 15.7 & 13.8 & 11.4 & 8.9 & 6.5 \\
\midrule
\multicolumn{8}{@{}l}{\textit{Referring, referring model with its grounding head}} \\
HumanRef & \multicolumn{7}{c}{59.0} \\
RefCOCOg & \multicolumn{7}{c}{88.7} \\
\bottomrule
\end{tabular}
\end{table}

\begin{table*}[t]
\centering
\footnotesize
\caption{F1@0.5 of \method on LA in the slow mode with coordinate anchors and with image-feature anchors, after SFT and after RL, the means of F1@0.5, precision, recall and F1@mIoU over the 12 box benchmarks, and the mean time per image.}
\label{tab:feat_anchor}
\setlength{\tabcolsep}{3.4pt}
\begin{tabular}{@{}lcccccccccccc|ccccc@{}}
\toprule
Anchor & Dense & COCO & VisD. & LVIS & HierT. & TotalT. & SROIE & IC15 & DocL. & M6Doc & HRef & RefG & F1 & P & R & mIoU & s/img \\
\midrule
\multicolumn{18}{@{}l}{\textit{SFT}} \\
Coordinates & 80.7 & 73.4 & 60.1 & 66.0 & 78.0 & 68.6 & 87.0 & 47.2 & 90.6 & 87.1 & 79.8 & 85.4 & 75.3 & 70.6 & 84.2 & 59.9 & 0.72 \\
Image feature & 71.6 & 74.5 & 58.7 & 67.2 & 76.2 & 70.4 & 81.6 & 50.2 & 90.2 & 88.5 & 81.2 & 85.9 & 74.7 & 73.4 & 79.6 & 59.6 & 0.70 \\
\midrule
\multicolumn{18}{@{}l}{\textit{SFT + RL}} \\
Coordinates & 81.0 & 74.7 & 62.5 & 66.7 & 78.8 & 71.3 & 89.6 & 50.3 & 91.5 & 89.0 & 82.7 & 86.7 & 77.1 & 72.6 & 84.9 & 61.4 & 0.71 \\
Image feature & 78.4 & 75.6 & 61.6 & 67.8 & 79.3 & 71.5 & 88.1 & 50.9 & 89.3 & 89.6 & 81.2 & 86.6 & 76.7 & 75.2 & 80.9 & 61.0 & 0.68 \\
\bottomrule
\end{tabular}
\end{table*}

\subsection{Evaluation of UFO}
\label{sec:ufo_eval}
We run the released UFO-InternVL2.5-8B models with their official code on the same samples as \cref{tab:main}.
The multi-task model, trained for COCO detection and segmentation at a resolution of 1344, resizes each image to $1344\times1344$, splits it into nine $448\times448$ tiles with $8\times8$ grid points each, keeps the three most probable categories of each of the 576 points, decodes the 300 most probable of these candidates in parallel for up to 30 tokens and applies soft-NMS within each category at IoU $0.5$.
The score of a box is the probability of its first token, which names its category.
On COCO its category list holds all 80 categories, as in its own evaluation, and we keep the boxes of the categories queried for the image.
On the other sets the list holds the queried categories.
On VisDrone, listing all ten VisDrone categories instead gives 50.6 F1, against 51.1 with the queried categories.
Its released configuration keeps up to 300 boxes per image without a score threshold, since the AP of its own evaluation ranks the boxes by score.
F1 counts a fixed set of boxes and needs a score threshold, so \cref{tab:main} keeps the boxes scored at least $0.5$, a threshold fixed before evaluation, and \cref{tab:ufo_sweep} reports F1 at score thresholds from $0.1$ to $0.7$.
On the text and layout sets it predicts none of the queried categories at any score threshold.
The referring model, fine-tuned at a resolution of 896 on referring and VQA data, decodes one box per expression with its official grounding head, which gives no score and needs no threshold.
Its grid mode with the expression as the only category finds few targets, even when any answer of a point other than background is taken as the expression.
Each sample is timed alone on one H100 GPU, including image loading, the pipeline and post-processing.

\subsection{Differences from \method}
\label{sec:ufo_diff}
Both decode the point queries of an image in parallel under an attention mask that keeps them apart, and in both a point can answer that it holds no object.
How a point enters the model is not a difference either, since \method also works when its anchors are given as image features, as in UFO (\cref{sec:feat_anchor}).
They differ in what they start from, in which queries they decode in parallel and in how a point is tied to an instance.

\textbf{A new model or a general strategy.}
UFO adds grid tokens whose local image feature is added again at every layer, makes attention among image tokens bidirectional, defines its own output of a category name and a box relative to the point, and trains the changed model on its own multi-task data.
\method is a strategy for VLMs that ground through their text output, such as the three bases of \cref{tab:main}, which write coordinates in different formats.
It changes only the prompt and the attention mask, keeps the native output format of the base and needs a post-training on 47{,}170 pairs, and each base is compared with its own sequential decoding trained on the same pairs.

\textbf{Which queries are decoded in parallel.}
The parallel path of UFO serves detection and segmentation over a category list given in the prompt, and a referring expression goes to a separate model that decodes one box in sequence (\cref{tab:ufo_sweep}).
In \method the anchor follows the query in an ordinary prompt, so the queries of the base, here objects, text lines, layout regions and referring expressions, are all decoded in parallel with any number of targets (\cref{tab:main}).

\textbf{Which instance a point reports.}
UFO assigns objects to grid points by Hungarian matching during training and decodes one fixed grid of 576 points for all categories.
\method states the assignment as a rule in the query, the matching instance whose center is nearest to the anchor within a radius, or \none.
With this rule the answer of a coarse anchor decides whether its cell is refined, so each query decodes 100 coarse anchors and 4 fine anchors in each occupied cell, and the anchors follow the density of the image.
Within \method, a single fine grid over the whole image takes 1.14 instead of 0.71 seconds per image, with lower recall and F1 (\cref{tab:abl_decoder}).
The rule also gives every anchor its own target, which lets the set-level RL credit each anchor with its part of the set F1.

\subsection{Local Image Features as Anchors}
\label{sec:feat_anchor}
Parallel streams of one query need an anchor that sets them apart, and \method sets them apart by position, writing a point into each query (\cref{sec:queries}).
Such an anchor can also be given by the local image feature at its point, as in UFO~\cite{tang2026ufo}, whose other differences from \method are described in \cref{sec:ufo_diff}.
We test this on LA by replacing the two coordinate tokens of every anchor with one placeholder token whose input embedding is this feature, interpolated bilinearly from the visual tokens that enter the language model.
Unlike UFO, we add the feature only at the input and not again at every layer, keep the causal attention of LA among image tokens, and keep the label of each anchor under the nearest-center rule and its answer in absolute coordinates.
Only the form of the anchor changes, and the pairs, the prefix tree, the grids, the training recipe, the decoding and the evaluation are those of \method.

The two kinds of anchors perform almost the same (\cref{tab:feat_anchor}).
After SFT and after RL, the feature anchors reach 74.7 and 76.7 F1 against 75.3 and 77.1 for coordinate anchors, with higher precision and lower recall.
Dense200, where many instances lie close together, shows the largest gap for the feature anchors, $-9.1$ F1 after SFT and $-2.6$ F1 after RL.
Timed as in \cref{tab:main}, the feature anchors take 0.70 and 0.68 seconds per image after SFT and after RL, against 0.72 and 0.71 for coordinate anchors.

\section{Qualitative Examples}
\label{sec:examples}
\begin{figure}[!t]
\centering
\includegraphics[width=\linewidth]{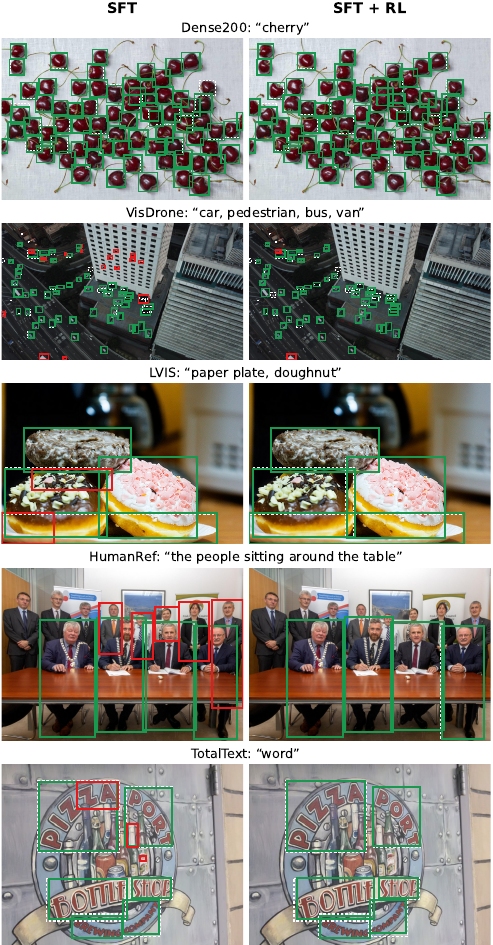}
\caption{Boxes of \method on LA after SFT (left) and after RL (right). Dashed white boxes are the ground truth, and green and red boxes are correct and false predictions at IoU $0.5$.}
\label{fig:examples}
\end{figure}

\Cref{fig:examples} shows the boxes of \method on LA in the slow mode on images of five box benchmarks, after SFT and after RL.
\method finds all 74 cherries on Dense200, most of the small vehicles in the streets around two tall buildings on VisDrone and the doughnuts and their plate on LVIS.
On HumanRef it boxes only the four people sitting at the table, and on TotalText it finds every word of a painted sign, including the curved ones.
After SFT, more boxes fall on the background or on instances that the query does not ask for, such as the windows of a tall building on VisDrone and the people standing behind the table on HumanRef, and RL removes most of them.

\end{document}